\documentclass[10pt,twocolumn,letterpaper]{article}

\usepackage{cvpr}              % To produce the CAMERA-READY version
\definecolor{cvprblue}{rgb}{0.21,0.49,0.74}
\usepackage[pagebackref,breaklinks,colorlinks,allcolors=cvprblue]{hyperref}
\usepackage{algorithm}
\usepackage{algorithmic}

\def\paperID{46042} % *** Enter the Paper ID here
\def\confName{CVPR}
\def\confYear{2026}

\title{SEBA: Sample-Efficient Black-Box Attacks on Visual Reinforcement Learning}

\author{
Tairan Huang\thanks{Backup email: htr.story@gmail.com; WeChat: trgg666.}\quad Yulin Jin \quad Junxu Liu \quad Qingqing Ye \quad Haibo Hu\thanks{Corresponding author.}
\\
The Hong Kong Polytechnic University\\
Hong Kong, China\\
{\tt\small \{tairan.huang, yulin.jin\}@connect.polyu.hk}
\\
{\tt\small \{junxu.liu, qqing.ye, haibo.hu\}@polyu.edu.hk}
}

\begin{document}
\maketitle
\begin{abstract}
Visual reinforcement learning has achieved remarkable progress in visual control and robotics, 
but its vulnerability to adversarial perturbations remains underexplored. 
Most existing black-box attacks focus on vector-based or discrete-action RL, and their effectiveness on image-based continuous control is limited by the large action space and excessive environment queries. 
We propose \textbf{SEBA}, a sample-efficient framework for black-box adversarial attacks on visual RL agents. 
SEBA integrates a shadow Q model that estimates cumulative rewards under adversarial conditions,
a generative adversarial network that produces visually imperceptible perturbations, 
and a world model that simulates environment dynamics to reduce real-world queries. 
Through a two-stage iterative training procedure that alternates between learning the shadow model and refining the generator, SEBA achieves strong attack performance while maintaining efficiency. 
Experiments on MuJoCo and Atari benchmarks show that SEBA significantly reduces cumulative rewards, preserves visual fidelity, and greatly decreases environment interactions compared to prior black-box and white-box methods.
The code is available at \href{https://github.com/tairanhuang/seba}{https://github.com/tairanhuang/seba} online.
\end{abstract}
    
\section{Introduction}
\label{sec:intro}

Reinforcement learning (RL) from visual observations~\cite{DBLP:conf/cvpr/XuPTLX024, DBLP:conf/cvpr/ChoiLSJSM23} has become a cornerstone of 
embodied intelligence, powering applications in robotic manipulation~\cite{manipulation}, autonomous navigation~\cite{navigation}, and visual control~\cite{control}.
Building on the foundation of deep reinforcement learning~\cite{origin_1, origin_2, cgar}, 
visual RL enables agents to learn control policies directly from raw pixel inputs 
and operate effectively in high-dimensional, dynamic environments without relying on hand-crafted state representations. 
However, this reliance on visual perception also makes such systems vulnerable to small, carefully crafted perturbations known as \textit{adversarial attacks}~\cite{DBLP:conf/aaai/Korkmaz23, DBLP:conf/iclr/HuangPGDA17, DBLP:conf/aaai/Korkmaz22, paad, korkmaz2020nesterov, atla, redefine}, 
which can cause severe degradation or even catastrophic failures in control and decision-making. 
Understanding and improving the robustness of visual RL agents is therefore critical for ensuring the safe deployment of learning-based autonomous systems in real-world environments.

Despite the growing attention to adversarial robustness in RL, existing studies remain limited in scope. Most prior works focus on \textit{vector-based} RL~\cite{sa-mdp, atla, paad} or \textit{discrete-action} visual RL (e.g., Atari)~\cite{DBLP:conf/aaai/Korkmaz22, redefine, sa-mdp, paad}, where the environment state is either low-dimensional or where the policy resembles a classifier~\cite{atla}. In contrast, \textit{continuous-action} visual RL presents a far more challenging setting: the action space is infinite, the observation space is high-dimensional, and perturbations can affect long-horizon dynamics in complex ways. 
To date, there has been little exploration of how to perform effective and efficient adversarial attacks in this setting, 
especially under the practical constraint of \textbf{black-box access}, 
where the attacker cannot observe gradients or internal model parameters.

Designing such an attack introduces multiple challenges.  
First, the black-box constraint prevents direct gradient computation, requiring the attacker to infer vulnerabilities purely from external observations of state, action, and reward.  
Second, naively estimating attack gradients via repeated environment queries is prohibitively expensive, 
as each rollout in RL entails costly sequential interaction.  
Third, the attacker must balance three competing objectives: attack \textit{strength} (reward degradation), \textit{imperceptibility} (maintaining visual realism), and \textit{sample efficiency} (minimizing query cost). 
Achieving all three simultaneously demands a principled framework that can model long-term reward effects, produce realistic perturbations, and reduce dependence on real environment feedback.

To address these challenges, we propose \textbf{SEBA} (\textit{Sample-Efficient Black-box Attacks}), a novel framework 
for efficient adversarial perturbation generation in visual RL with black-box access. 
SEBA integrates four key components:
a \textit{shadow Q model} that estimates cumulative reward under perturbations, 
a \textit{GAN module} for perceptual realism,
a \textit{two-stage alternating training scheme} to stabilize optimization, 
and a \textit{world model module} that improves sample efficiency via synthetic rollouts.

We conduct extensive experiments to validate SEBA.  
On \textbf{MuJoCo} visual control tasks, SEBA surpasses all existing white-box and black-box baselines, achieving state-of-the-art performance in terms of reward degradation, visual imperceptibility, and query efficiency.  
On \textbf{Atari} benchmarks, SEBA remains highly competitive, outperforming most white-box methods and second only to PA-AD, while maintaining substantially lower query costs.  
We further demonstrate SEBA’s ability to perform \emph{targeted} attacks, precisely steering specific action dimensions toward desired outcomes.  
These results collectively highlight SEBA’s 
effectiveness, efficiency, and subtlety 
across diverse reinforcement learning domains.

\noindent\textbf{Contributions.}  
This work makes the following key contributions:
\begin{itemize}
    \item We tackle the open challenge of performing \textit{sample-efficient black-box adversarial attacks} on visual reinforcement learning agents, where direct gradient access and extensive environment interaction are both infeasible.
    \item We propose \textbf{SEBA}, a two-stage framework that combines a shadow $Q$ model, a GAN-based perturbation generator, and a learned world model. 
    The shadow $Q$ model supplies $Q$-based guidance without accessing the victim’s internals, while the world model generates synthetic rollouts to reduce real-environment queries.
    \item Comprehensive experiments across both continuous-control (\textbf{MuJoCo}) and discrete-action (\textbf{Atari}) domains show that SEBA achieves stronger attack effectiveness, higher visual imperceptibility, and greater query efficiency than existing white-box and black-box methods.
\end{itemize}

SEBA offers a principled and practical framework for efficient visual attacks in reinforcement learning, advancing the state of the art toward realistic, deployable black-box adversarial evaluation in embodied AI systems.

\section{Related Work}
\subsection{Visual Reinforcement Learning}
Visual RL learns policies directly from raw images, requiring representation learning to encode observations into compact latent features. Existing methods are broadly categorized into prediction-based and image-based approaches. Prediction-based methods~\cite{DBLP:conf/icml/ZhangRXS024,dreamer} learn latent dynamics to facilitate training.
For example,
IRIS~\cite{iris} combines a discrete autoencoder with an autoregressive Transformer to learn a world model, ~\cite{DBLP:conf/icml/TangGRPCMRALL0T23} proposes bidirectional self-predictive learning, and DeepMDP~\cite{deepmdp} learns latent representations by predicting rewards and future latent states. 
In contrast, image-based methods leverage vision techniques for encoder learning: DrQ~\cite{drq} learns encoders via data augmentation without auxiliary tasks, CURL~\cite{curl} adopts contrastive learning with query-key pairs, 
SAC-AE~\cite{sacae} enhances training with deterministic autoencoders and reconstruction loss,
and A-LIX~\cite{alix} mixes latent components with neighbors in the feature map.

\subsection{Adversarial Attack on Reinforcement Learning}
Adversarial attacks expose the fragility of learning-based decision systems~\cite{lironghua, xiaoyaxin, liangzi, jin1, jin2}.
Adversarial attacks in RL craft perturbations to mislead agents and degrade performance.
Prior work explores various attack strategies.
\cite{DBLP:conf/iclr/HuangPGDA17, DBLP:conf/atal/PattanaikTLBC18} applies gradient-based methods to manipulate agent behavior. 
\cite{korkmaz2020nesterov} introduces a Nesterov-momentum–based optimization method for crafting adversarial perturbations in deep RL.
SA-MDP~\cite{sa-mdp} proposes critic-based and maximal-action-difference attacks. 
OPTIMAL~\cite{atla} defines the adversary’s reward as the negative of the victim’s return, allowing the attacker to learn through reinforcement learning. 
PA-AD~\cite{paad} combines an RL-based director with a non-RL actor, forming a white-box attack. 
\cite{DBLP:conf/iclr/LiuGCZT0Z23} introduces Maximum Cost and Maximum Reward attackers. 
SAIA~\cite{saia} employs imitation learning for targeted attacks, while \cite{metamdp} adopts a meta-MDP framework to model interactions between the attacker, victim, and environment.

Most existing attacks focus on agents with vector-based states. 
Although SA-MDP~\cite{sa-mdp}, PA-AD~\cite{paad}, \cite{redefine} and \cite{DBLP:conf/aaai/Korkmaz22} have been extended to visual RL, they are limited to discrete-action Atari environments, where the policy network acts like a classifier~\cite{atla}. 
To our knowledge, this is the first work to investigate visual RL attacks in continuous control tasks, a more challenging setting.

\section{Method: SEBA}
\label{sec:method}

\begin{figure*}[t]
  \centering
  \includegraphics[width=0.98\linewidth]{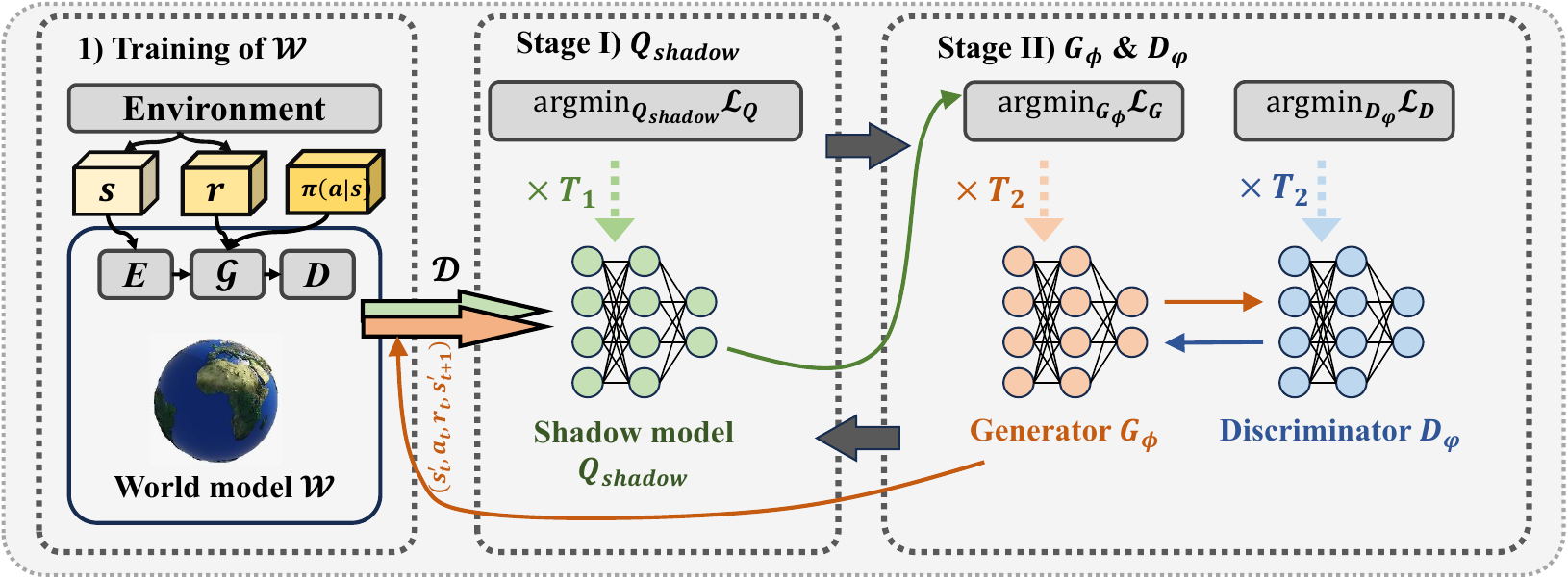}
  \caption{\textbf{SEBA overview.} The world model is first trained to predict visual dynamics and generate synthetic rollouts. 
  SEBA then alternates between updating the shadow critic on perturbed states
  and optimizing the GAN.}
  \label{fig:seba_overview}
\end{figure*}

In this section, we present the proposed \textbf{SEBA} (Sample-Efficient Black-box Attack) framework for visual reinforcement learning. 
We first describe the overall problem formulation and adversarial setting (\S\ref{sec:problem}). 
Then, we introduce the \emph{shadow Q model} that provides black-box guidance (\S\ref{sec:shadowq}) 
and the GAN-based perturbation module for generating imperceptible yet effective attacks (\S\ref{sec:gan}). 
Next, we detail the two-stage alternating optimization strategy (\S\ref{sec:twostage}) 
and the incorporation of a learned world model to improve query efficiency (\S\ref{sec:wm}). 
Finally, we summarize the overall training pipeline and algorithms (\S\ref{sec:overall}).
An overview of the complete procedure is shown in Fig.~\ref{fig:seba_overview}.

\subsection{Problem Setup}
\label{sec:problem}
Let $\mathcal{M}=\langle\mathcal{S},\mathcal{A},p,r,\gamma\rangle$ be an MDP with image observations $s\in[0,1]^{C\times H\times W}$ and continuous actions $a\in\mathcal{A}\subset\mathbb{R}^d$. A victim policy $\pi(a\mid s)$ is given in \emph{black-box} form: we may query $\pi$ to obtain actions but cannot access internals or gradients. The attacker outputs a bounded perturbation $\delta=\mathrm{clip}(G_\phi(s),-\epsilon,\epsilon)$ and forms $s'=\mathrm{clip}(s+\delta,0,1)$. The attack objective is to minimize the discounted return of the victim:
\begin{equation}
\min_{\phi}\;\mathbb{E}\!\left[\sum_{t=0}^{\infty}\gamma^t\,r\big(s'_t,a_t\big)\right],\quad
a_t\sim \pi(\cdot\mid s'_t).
\label{eq:attack_obj}
\end{equation}

\subsection{Shadow Model Training}
\label{sec:shadowq}

The shadow critic \(Q_{\text{shadow}}\) serves as a differentiable surrogate for the victim policy, estimating its expected cumulative reward under adversarial perturbations. This model provides the attacker's optimization signal in a fully black-box setting.

During training, the generator \(G_\phi\) adds bounded perturbations \(\delta_t\) to clean observations \(s_t\) to form adversarial states \(s'_t = s_t + \delta_t\). 
The victim policy \(\pi(a|s'_t)\) is then executed to produce the corresponding action \(a_t\),
which interact with the environment to produce rewards \(r_t\) and next states \(s_{t+1}\). The process repeats to yield transition tuples \((s'_t, a_t, r_t, s'_{t+1}, \texttt{done}_t)\) stored in a replay buffer for updating \(Q_{\text{shadow}}\).

The shadow model is optimized via the temporal-difference (TD) objective:
\begin{equation}
\begin{aligned}
\mathcal{L}_Q
&= \tfrac{1}{2}\,\mathbb{E}\!\left[
  \big(Q_{\text{shadow}}(s'_t, a_t) - y_t \big)^2
\right],\\
y_t
&= r_t + \gamma\,
\mathbb{E}_{a\sim\pi(\cdot\mid s'_{t+1})}
\!\left[ Q_{\text{shadow}}(s'_{t+1}, a) \right],
\label{eq:td1}
\end{aligned}
\end{equation}
where \(r_t\) is the immediate reward and \(\gamma\) is the discount factor.  
The victim policy \(\pi\) is executed only to produce actions, without accessing its gradients or internal parameters.

To enhance training efficiency, SEBA alternates between \textbf{real-environment updates} and \textbf{world-model updates}.  
During $Q_{\text{shadow}}$ training, both real and synthetic transitions are utilized, ensuring accurate supervision from real rollouts while leveraging the world model to reduce interaction cost.

\subsection{GAN-based Perturbation Generation}

\label{sec:gan}

To generate effective yet visually imperceptible perturbations, SEBA employs a generative adversarial framework composed of a generator $G_\phi$ and a discriminator $D_\psi$. 
The generator produces bounded pixel perturbations, while the discriminator enforces realism by distinguishing clean states from their adversarial counterparts.

Given a batch of observations $\{s_k\}_{k=1}^B$, adversarial states are formed as 
$s'_k=\mathrm{clip}(s_k+G_\phi(s_k),0,1)$. 
The victim policy $\pi(a|s'_k)$ is then executed to obtain corresponding actions $a_k$. 
These interactions can obtained from either the real environment or the learned world model (Sec.~\ref{sec:wm}), depending on whether training uses real or simulated experience.

The discriminator and generator are optimized with opposing objectives:
\begin{equation}
\begin{aligned}
\mathcal{L}_D &= -\tfrac{1}{B}\!\sum_{k=1}^B [\log D_\psi(s_k)+\log(1-D_\psi(s'_k))],\\
\mathcal{L}_G &= -\tfrac{1}{B}\!\sum_{k=1}^B \!\big(\log D_\psi(s'_k)-\lambda Q_{\text{shadow}}(s'_k,a_k)\big).
\label{eq:gan_losses}
\end{aligned}
\end{equation}

The first term in $\mathcal{L}_G$ enforces imperceptibility, while the second drives reward minimization via the shadow critic under a black-box constraint.

\subsection{Two-Stage Alternating Optimization}
\label{sec:twostage}
Directly optimizing the generator $G_\phi$ and the shadow critic $Q_{\text{shadow}}$ together often leads to instability, as their objectives are tightly coupled.  
To stabilize learning, SEBA adopts a two-stage alternating strategy that separates the updates of the critic and the generator:

\noindent\textbf{Stage 1 (Shadow Model Update).}  
Freeze $(G_\phi, D_\psi)$, roll out adversarial interactions to collect tuples $(s', a, r, s'_{+1})$,  
and update $Q_{\text{shadow}}$ using the temporal-difference loss in Eq.~\eqref{eq:td1}, ensuring that it accurately models the victim’s reward dynamics under perturbations.

\noindent\textbf{Stage 2 (GAN Update).}  
Freeze $Q_{\text{shadow}}$, generate adversarial states $s'$, obtain corresponding actions $a$ from the victim policy, and update $(G_\phi, D_\psi)$ according to Eq.~\eqref{eq:gan_losses}.  
This step optimizes the generator for both attack effectiveness and imperceptibility, guided by the stable supervision from the fixed critic.

\subsection{World Model for Query Efficiency}
\label{sec:wm}

To alleviate the cost of querying the real environment, 
SEBA employs a \emph{world model} $\mathcal{W}$ trained from the replay buffer $\mathcal{D}$.
Following the IRIS framework~\citep{iris}, $\mathcal{W}$ consists of a discrete image tokenizer $(E, D)$ and an autoregressive Transformer $\mathcal{G}$ that predicts future latent tokens and rewards conditioned on past state--action sequences:
\begin{align}
z_t &= E(s_t), \\
\hat{z}_{t+1},\, \hat{r}_t &= \mathcal{G}(z_{\le t}, a_{\le t}), \\
\hat{s}_{t+1} &= D(\hat{z}_{t+1}).
\label{eq:wm_dynamics}
\end{align}

\paragraph{World-model training.}
$\mathcal{W}$ is trained on samples from replay buffer $\mathcal{D}$ to predict next latent tokens and rewards. 
We use a single joint prediction loss:
\begin{equation}
\begin{aligned}
\mathcal{L}_{\mathcal{W}} =\;
&\mathbb{E}_{(s_t,a_t,r_t,s_{t+1})\sim\mathcal{D}}
\left[-\log p_{\mathcal{G}}(z_{t+1}\mid z_{\le t},a_{\le t})\right] \\
&+
\mathbb{E}_{(s_t,r_t)\sim\mathcal{D}}
\left[\|\hat r_t - r_t\|_2^2\right].
\end{aligned}
\label{eq:wm_loss}
\end{equation}

After $N_{\mathcal{W}}$ training steps, $\mathcal{W}$ is used to generate imagined transitions for SEBA training.

Once trained via Eq.~\ref{eq:wm_loss}, the world model captures visual and reward dynamics, enabling the attacker to generate synthetic transitions $(\hat{s}'_t, a_t, \hat{r}_t, \hat{s}'_{t+1})$ without invoking the real environment.

\paragraph{Model-based rollouts.}
After training, the world model $\mathcal{W}$ is employed to generate \emph{model rollouts}, which simulate transitions for updating both the shadow critic $Q_{\text{shadow}}$ and the generator $G_\phi$:
\begin{equation}
\begin{aligned}
y_t &= \hat{r}_t + \gamma\,
\mathbb{E}_{a\sim\pi(\cdot\mid \hat{s}'_{t+1})}
\!\left[ Q_{\text{shadow}}(\hat{s}'_{t+1}, a) \right],
\label{eq:wm_td}\\[4pt]
\end{aligned}
\end{equation}
\begin{equation}
\begin{aligned}
\mathcal{L}_G &= -\tfrac{1}{B}\!\sum_{k=1}^B \!\big(\log D_\psi(\hat{s}'_k)-\lambda Q_{\text{shadow}}(\hat{s}'_k,a_k)\big).
\label{eq:wm_gan}
\end{aligned}
\end{equation}

Each real interaction is paired with $H$ model-generated transitions, 
effectively expanding the replay buffer and reducing real environment queries by $\mathcal{O}(H)$.

\begin{algorithm}[t]
\caption{SEBA: Sample-Efficient Black-box Attack}
\label{alg:seba_overall}
\begin{algorithmic}[1]

\STATE Initialize $G_\phi, D_\psi, Q_{\text{shadow}}, \mathcal{W}$ and buffer $\mathcal{D}\!\leftarrow\!\emptyset$

\STATE Collect real transitions $(s,a,r,s_{+1})$ and train $\mathcal{W}$ by minimizing $\mathcal{L}_{\mathcal{W}}$ (Eq.~\ref{eq:wm_loss})

\FOR{$i=1$ to $N_{\mathrm{iter}}$}
    \STATE Update $Q_{\text{shadow}}$ (Stage~1, Alg.~\ref{alg:stage1}) with TD loss
    \STATE Update $(G_\phi,D_\psi)$ (Stage~2, Alg.~\ref{alg:stage2}) with fixed $Q_{\text{shadow}}$
\ENDFOR

\STATE \textbf{return} $G_\phi$
\end{algorithmic}
\end{algorithm}

\subsection{Overall procedure}
\label{sec:overall}

Algorithm~\ref{alg:seba_overall} summarizes SEBA.
The world model $\mathcal{W}$ generates synthetic rollouts that are mixed with real transitions in $\mathcal{D}$ for joint updates of $Q_{\text{shadow}}$ and $G_\phi$.  
Periodic real-environment queries are interleaved to correct model drift and maintain fidelity.  
SEBA reuses $H$ synthetic transitions per real step, reducing environment queries by about $1/H$.  
Final evaluations are performed in the real environment.

\begin{algorithm}[t]
\caption{Stage 1: Train Shadow Critic $Q_{\text{shadow}}$}
\label{alg:stage1}
\begin{algorithmic}[1]
\STATE // Shadow-critic update loop
\FOR{$t=1$ to $T_1$}
  \STATE Sample $s_t$ from environment
  \STATE Compute $s'_t \!\leftarrow\! \operatorname{clip}\!\big(s_t+\operatorname{clip}(G_\phi(s_t),-\epsilon,\epsilon),0,1\big)$
  \STATE Obtain action $a_t \!\sim\! \pi(\cdot|s'_t)$
  \STATE Step environment to get $(r_t, s_{t+1}, done_t)$
  \IF{use\_wm}
    \STATE Roll out $\mathcal{W}$ for $H$ steps from $s_t$ to produce synthetic transitions for $\mathcal{D}$
  \ELSE
    \STATE Append $(s'_t,a_t,r_t,s'_{t+1},done_t)$ to $\mathcal{D}$
  \ENDIF
  \STATE Sample a minibatch from $\mathcal{D}$;
  \STATE update $Q_{\text{shadow}}$ by the loss in Eq.~\eqref{eq:td1}
\ENDFOR
\end{algorithmic}
\end{algorithm}

\begin{algorithm}[t]
\caption{Stage 2: Train $(G_\phi, D_\psi)$}
\label{alg:stage2}
\begin{algorithmic}[1]
\STATE // Generator–discriminator update loop
\FOR{$t=1$ to $T_2$}
  \STATE Sample $s_t$ from environment or world model $\mathcal{W}$
  \STATE Compute $s'_t \!\leftarrow\! \operatorname{clip}\!\big(s_t+\operatorname{clip}(G_\phi(s_t),-\epsilon,\epsilon),0,1\big)$
  \STATE Obtain action $a_t \!\sim\! \pi(\cdot|s'_t)$
  \STATE Update $(G_\phi, D_\psi)$ using Eq.~\eqref{eq:gan_losses}
\ENDFOR
\end{algorithmic}
\end{algorithm}

\section{Experiments}
\label{sec:experiments}

In this section, we present a comprehensive evaluation of SEBA. 
We consider four aspects: 
(i) pixel-based MuJoCo control tasks with DrQ-SAC victims, 
comparing against both 2D image-space attack algorithms and vector-state RL attack algorithms, 
evaluated in terms of effectiveness (reward), imperceptibility (FID), and query efficiency (query cost);
(ii) transfer to Atari with PPO agents to assess cross-domain generality; 
(iii) ablations of key components, including the discriminator, the use of noisy states in Stage~1, and the world model;
and (iv) targeted attacks on selected action dimensions for fine-grained behavioral control.

\begin{table*}[!t]
\centering
\caption{\textbf{Image-based attacks on pixel DrQ-SAC agents.}
Lower reward and FID indicate stronger attacks. Query cost includes training-time interactions with the environment or victim agent, and attack-time victim queries per step.}
\label{tab:image_attacks}
\resizebox{\textwidth}{!}{
\begin{tabular}{lcccccc}
\toprule
& \textbf{No Attack} & \multicolumn{2}{c}{\textbf{White-box}} & \multicolumn{3}{c}{\textbf{Black-box}} \\
\cmidrule(lr){2-2}\cmidrule(lr){3-4}\cmidrule(lr){5-7}
\textbf{Task (Reward $\downarrow$)} & Clean & PGD & C\&W & SimBA & Square & \textbf{SEBA} \\
\midrule
Cheetah Run   & $859.26_{\,\pm\,35.00}$ & $150.72_{\,\pm\,49.52}$ & $183.41_{\,\pm\,50.01}$ & $52.15_{\,\pm\,20.09}$ & $182.90_{\,\pm\,23.09}$ & $\mathbf{1.61_{\,\pm\,3.97}}$ \\
Walker Walk   & $944.28_{\,\pm\,42.26}$ & $342.78_{\,\pm\,35.82}$ & $438.67_{\,\pm\,28.56}$ & $68.77_{\,\pm\,15.44}$ & $752.19_{\,\pm\,22.22}$ & $\mathbf{35.74_{\,\pm\,6.67}}$ \\
Walker Run    & $718.15_{\,\pm\,26.79}$ & $118.35_{\,\pm\,16.38}$ & $134.43_{\,\pm\,25.92}$ & $32.87_{\,\pm\,5.28}$ & $258.41_{\,\pm\,16.31}$ & $\mathbf{17.09_{\,\pm\,4.73}}$ \\
Reacher Hard  & $870.9_{\,\pm\,90.43}$ & $232.3_{\,\pm\,31.71}$ & $332.0_{\,\pm\,40.84}$ & $2.7_{\,\pm\,6.60}$ & $870.7_{\,\pm\,90.38}$ & $\mathbf{0.3_{\,\pm\,0.9}}$ \\
Hopper Stand  & $849.60_{\,\pm\,128.49}$ & $1.85_{\,\pm\,5.56}$ & $8.22_{\,\pm\,14.61}$ & $4.86_{\,\pm\,6.72}$ & $652.10_{\,\pm\,86.02}$ & $\mathbf{1.25_{\,\pm\,3.75}}$ \\
\midrule
FID $\downarrow$ & / & 109.43 & 110.97 & 78.05 & 118.01 & \textbf{62.43} \\
\midrule
Train Env (total) $\downarrow$    & / & / & / & / & / & 160K \\
Train Vic (total) $\downarrow$ & / & / & / & / & / & 800K \\
Atk. Vic (per-step) $\downarrow$ & / & 20 & 20 & 400 & 202 & 0 \\
\bottomrule
\end{tabular}}
\end{table*}

\begin{table*}[!t]
\centering
\caption{\textbf{Vector-state RL attacks adapted to pixel-based control.}
Methods originally designed for low-dimensional vector observations degrade in high-dimensional image space, 
while SEBA remains both effective and query-efficient.}
\label{tab:vector_attacks}
\resizebox{\textwidth}{!}{
\begin{tabular}{lcccccc}
\toprule
& \textbf{No Attack} & \multicolumn{3}{c}{\textbf{White-box}} & \multicolumn{2}{c}{\textbf{Black-box}} \\
\cmidrule(lr){2-2}\cmidrule(lr){3-5}\cmidrule(lr){6-7}
\textbf{Task (Reward $\downarrow$)} & Clean & Critic-Based & MAD & PA-AD & OPTIMAL & \textbf{SEBA} \\
\midrule
Cheetah Run   & $859.26_{\,\pm\,35.00}$ & $117.86_{\,\pm\,39.44}$ & $29.02_{\,\pm\,11.42}$ & $146.61_{\,\pm\,24.44}$ & $271.73_{\,\pm\,34.72}$ & $\mathbf{1.61_{\,\pm\,3.97}}$ \\
Walker Walk   & $944.28_{\,\pm\,42.26}$ & $231.03_{\,\pm\,99.92}$ & $149.58_{\,\pm\,43.54}$ & $323.96_{\,\pm\,51.72}$ & $631.30_{\,\pm\,138.28}$ & $\mathbf{35.74_{\,\pm\,6.67}}$ \\
Walker Run    & $718.15_{\,\pm\,26.79}$ & $117.42_{\,\pm\,16.26}$ & $38.63_{\,\pm\,4.58}$ & $186.15_{\,\pm\,13.54}$ & $325.86_{\,\pm\,15.69}$ & $\mathbf{17.09_{\,\pm\,4.73}}$ \\
Reacher Hard  & $870.9_{\,\pm\,90.43}$ & $213.6_{\,\pm\,72.99}$ & $26.19_{\,\pm\,16.5}$ & $45.37_{\,\pm\,28.44}$ & $592.64_{\,\pm\,116.13}$ & $\mathbf{0.3_{\,\pm\,0.9}}$ \\
Hopper Stand  & $849.60_{\,\pm\,128.49}$ & $8.50_{\,\pm\,17.11}$ & $1.32_{\,\pm\,3.97}$ & $25.02_{\,\pm\,5.99}$ & $331.91_{\,\pm\,68.83}$ & $\mathbf{1.25_{\,\pm\,3.75}}$ \\
\midrule
FID $\downarrow$ & / & 115.46 & 106.34 & 97.55 & 93.04 & \textbf{62.43} \\
\midrule
Train Env (total) $\downarrow$    & / & / & / & 4M & 4M & 160K \\
Train Vic (total) $\downarrow$ & / & / & / & 4M & 4M & 800K \\
Atk. Vic (per-step) $\downarrow$ & / & 20 & 11 & 0 & 0 & 0 \\
\bottomrule
\end{tabular}
}
\end{table*}

\subsection{Experimental Setup}

\noindent\textbf{Environments and victim agents.}
We evaluate SEBA on five pixel-based continuous-control tasks from the MuJoCo simulator, 
which provides diverse physics-based benchmarks for reinforcement learning.
Victim agents are trained with DrQ-SAC~\cite{sac,drq}, which extends SAC by incorporating data augmentation into the training process.
We adopt DrQ-SAC because it achieves strong performance without auxiliary losses and forms the backbone of several visual RL algorithms~\cite{gentle,pisac}.
All victim policies are pre-trained and fixed during attack evaluation.
Additional results on other visual RL attacks are reported in the Supplementary Material.

\noindent\textbf{Baselines.}
To our knowledge, we are the first to study \emph{black-box attacks on image-based continuous-control RL agents}, specifically in visual MuJoCo environments.
Accordingly, we compare SEBA against two families of baselines:

(i) \emph{2D image-space attacks} that directly perturb pixels (PGD~\cite{pgd}, C\&W~\cite{cw}, SimBA~\cite{simba}, Square~\cite{square});

(ii) \emph{vector-state RL attacker methods} originally developed for low-dimensional observations and adapted here to pixel control (Critic-Based~\cite{sa-mdp}, MAD~\cite{sa-mdp}, PA-AD~\cite{paad}, OPTIMAL~\cite{atla}).

All baselines are tuned to their best-reported configurations.
For fairness, PA-AD and OPTIMAL are adapted to pixel control by using the same generator architecture as SEBA.
Implementation details are provided in the Supplementary Material.

\noindent\textbf{Evaluation metrics.}
We report three metrics:

(1) \textbf{Effectiveness:} cumulative reward (lower is better).

(2) \textbf{Imperceptibility:} measured by the Fr\'echet Inception Distance (FID)~\cite{fid} between clean and adversarial state distributions (lower is better).

% \quad
(3) \textbf{Query efficiency:} we report \emph{Train Env (total)} for total environment queries during attacker training, 
\emph{Train Vic (total)} for total victim queries during attacker training, 
and \emph{Atk. Vic (per-step)} for victim queries per step during attack execution.
We omit per-step environment queries during attack execution because they are zero for all methods.

All results are averaged over 10 random seeds and reported as mean $\pm$ standard deviation.

\noindent\textbf{Model components.}
SEBA uses a shadow critic trained with a Double-DQN~\cite{ddqn} update on perturbed states.
The world model adopts an IRIS-style tokenizer, Transformer dynamics, and decoder to generate model-based rollouts.  
Implementation details are available in our code.

\noindent\textbf{Hyperparameters.}
SEBA and all baseline methods use the same perturbation bound $\epsilon=8/255$ for fairness, 
with $\lambda=1$ in the generator loss (Eq.~\ref{eq:gan_losses}). 
Key settings include world-model rollout horizon $H=4$, number of world-model updates $N_w=200$K, 
total training iterations $N_{\text{iter}}=20$, and phase lengths $T_1=T_2=5$K. 
Additional experiments with varying hyperparameters and other implementation details 
are reported in the Supplementary Material.

\subsection{Pixel-Based MuJoCo Results}

\paragraph{Overview.}
Tables~\ref{tab:image_attacks}--\ref{tab:vector_attacks} summarize the results across five pixel-based MuJoCo tasks. We assess attack performance in terms of reward degradation, perturbation imperceptibility, and query usage.

\paragraph{Effectiveness.}
We report the cumulative reward of the victim policy under attack.
SEBA yields the lowest rewards on all tasks. 
Representative examples: on \textit{Cheetah Run} SEBA obtains $\mathbf{1.61_{\pm3.97}}$ vs.\ PGD $150.72_{\pm49.52}$; on \textit{Reacher Hard} SEBA obtains $\mathbf{0.3_{\pm0.9}}$ vs.\ OPTIMAL $592.64_{\pm116.13}$.
These results show SEBA's consistent ability to induce long-horizon failures in image-based control.

\begin{figure}[t]
    \centering
    % === Row 1: Cheetah Run ===
    \begin{subfigure}{\linewidth}
        \centering
        \begin{subfigure}{0.19\linewidth}
            \includegraphics[width=\linewidth]{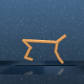}
            \caption*{Clean}
        \end{subfigure}
        \begin{subfigure}{0.19\linewidth}
            \includegraphics[width=\linewidth]{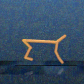}
            \caption*{PGD}
        \end{subfigure}
        \begin{subfigure}{0.19\linewidth}
            \includegraphics[width=\linewidth]{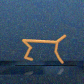}
            \caption*{C\&W}
        \end{subfigure}
        \begin{subfigure}{0.19\linewidth}
            \includegraphics[width=\linewidth]{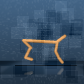}
            \caption*{Square}
        \end{subfigure}
        \begin{subfigure}{0.19\linewidth}
            \includegraphics[width=\linewidth]{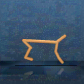}
            \caption*{SEBA}
        \end{subfigure}
        \caption*{Cheetah Run}
    \end{subfigure}
    \vspace{4pt}
    % === Row 2: Walker Run ===
    \begin{subfigure}{\linewidth}
        \centering
        \begin{subfigure}{0.19\linewidth}
            \includegraphics[width=\linewidth]{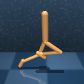}
            \caption*{Clean}
        \end{subfigure}
        \begin{subfigure}{0.19\linewidth}
            \includegraphics[width=\linewidth]{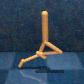}
            \caption*{PGD}
        \end{subfigure}
        \begin{subfigure}{0.19\linewidth}
            \includegraphics[width=\linewidth]{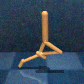}
            \caption*{C\&W}
        \end{subfigure}
        \begin{subfigure}{0.19\linewidth}
            \includegraphics[width=\linewidth]{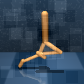}
            \caption*{Square}
        \end{subfigure}
        \begin{subfigure}{0.19\linewidth}
            \includegraphics[width=\linewidth]{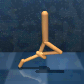}
            \caption*{SEBA}
        \end{subfigure}
        \caption*{Walker Run}
    \end{subfigure}

    \caption{\textbf{Visual comparison of perturbations.} SEBA produces subtle yet highly effective perturbations on pixel-based MuJoCo tasks. Each row shows clean and adversarial observations under different attack methods.}

    \label{fig:vis_mujoco}
\end{figure}

\paragraph{Imperceptibility.}
We assess visual subtlety using FID (lower is better).  
SEBA achieves the lowest FID across all tasks, indicating that its perturbations remain much closer to the clean observation distribution while still inducing strong performance degradation.  
Other baselines show noticeably higher FID, corresponding to larger visual deviation from clean states.  
Representative examples are shown in Fig.~\ref{fig:vis_mujoco}.

\paragraph{Query efficiency.}
SEBA performs attacks with \emph{zero} victim queries at execution time (\emph{Atk. Vic}=0),
whereas SimBA and Square require 400 and 202 queries per step, respectively (Table~\ref{tab:image_attacks}).
PA-AD and OPTIMAL incur substantially higher training cost, requiring roughly 4M \emph{Env} and \emph{Vic} queries during attacker training,
while SEBA uses 160K \emph{Env} and 800K \emph{Vic} queries.  
This highlights SEBA’s superior sample efficiency and its practicality when both environment interactions and victim queries are limited.

\paragraph{Focused comparison: PA-AD and OPTIMAL.}
PA-AD~\cite{paad} and OPTIMAL~\cite{atla} are the state-of-the-art attacker frameworks in the \emph{vector-state} setting: PA-AD is the strongest \emph{white-box} method, and OPTIMAL is the strongest \emph{black-box} RL-based attacker. 
Both methods treat the attacker as a policy \(\pi_a\) that generates perturbations \(\delta_t\) and optimize the long-horizon objective
\begin{equation}
    \min_{\pi_a}\; J(\pi_a)
    = \mathbb{E}\!\left[\sum_{t} -r_t(s_t + \delta_t)\right], 
    % \qquad 
    \delta_t \sim \pi_a(\cdot\mid s_t).
    \label{eq:paad_optimal_obj}
\end{equation}

This formulation is tractable when the observation space is low-dimensional. In pixel-based control, however, the perturbation lies in
\begin{equation}
    d = \mathrm{dim}(\delta_t) = 3 \times 84 \times 84 \approx 2 \times 10^{4},
    \label{eq:highdim}
\end{equation}
so the attacker must optimize in an \textbf{extremely high-dimensional action space}. 
This directly leads to: (i) a sharp increase in sample complexity, since exploration in \(\mathbb{R}^d\) becomes inefficient; and (ii) weak, noisy gradient signals, making policy optimization unstable or ineffective.

In contrast, SEBA does not learn a policy in this high-dimensional perturbation space. Instead, it directly trains a generator \(G_\phi\) guided by the shadow critic:
\begin{equation}
    \delta_t = G_\phi(s_t), 
    % \qquad 
    \nabla_\phi \mathcal{L}_G = -\,\nabla_\phi\, Q_{\text{shadow}}(s_t+\delta_t, a_t),
    \label{eq:seba_gen_update}
\end{equation}
so optimization focuses only on perturbations that directly reduce value, without requiring RL-based exploration over \(\mathbb{R}^d\). 
This allows SEBA to remain effective in pixel space while PA-AD and OPTIMAL degrade significantly (Table~\ref{tab:vector_attacks}).

\begin{table*}[t]
\centering
\caption{\textbf{Attacks on pixel-based Atari Rainbow agents.}
Lower reward and FID indicate stronger attacks; query rows report training-time ENV/VIC queries and attack-time victim queries per step.}
\label{tab:atari_attacks}
\begin{tabular}{lcccccc}
\toprule
& \textbf{No Attack} & \multicolumn{3}{c}{\textbf{White-box}} & \multicolumn{2}{c}{\textbf{Black-box}} \\
\cmidrule(lr){2-2}\cmidrule(lr){3-5}\cmidrule(lr){6-7}
\textbf{Task (Reward $\downarrow$)} & Clean & Critic-Based & MAD & PA-AD & OPTIMAL & SEBA \\
\midrule
Freeway     & $34_{\pm1}$     & $24_{\pm0}$     & $21_{\pm4}$     & $\mathbf{6_{\pm3}}$     & $32_{\pm5}$     & $\mathbf{10_{\pm4}}$ \\
Pong        & $21_{\pm0}$     & $14_{\pm4}$     & $15_{\pm1}$     & $\mathbf{-21_{\pm0}}$   & $18_{\pm2}$     & $\mathbf{3_{\pm0}}$ \\
Alien       & $8858_{\pm1007}$ & $3911_{\pm502}$ & $3364_{\pm698}$ & $\mathbf{443_{\pm281}}$ & $7400_{\pm1623}$ & $\mathbf{982_{\pm542}}$ \\
\midrule
FID $\downarrow$ & / & 131.6 & 136.4 & 102.4 & 113.9 & \textbf{81.7} \\
\midrule
Train Env (total) $\downarrow$    & / & / & / & 2M & 2M & 80K \\
Train Vic (total) $\downarrow$    & / & / & / & 2M & 2M & 400K \\
Atk. Vic (per-step) $\downarrow$  & / & 20 & 11 & 0 & 0 & 0 \\
\bottomrule
\end{tabular}
\end{table*}

\subsection{Atari Experiments}
Table~\ref{tab:atari_attacks} evaluates SEBA on three Atari games (Freeway, Pong, Alien) using \emph{Rainbow}~\cite{rainbow} as the victim agent. The state space remains pixel-based as in MuJoCo, but the action space in Atari is \emph{discrete} rather than continuous. This setting tests whether SEBA can still shape behavior when actions are selected from a finite action set.

SEBA consistently reduces reward across all three games. 
For example, on \textit{Freeway} the reward decreases from $34 \to 10$, and on \textit{Alien} from $8858 \to 982$. 
SEBA outperforms both white-box Critic-Based~\cite{sa-mdp} and MAD~\cite{sa-mdp}, as well as the black-box OPTIMAL~\cite{atla} baseline. 
Although \textbf{PA-AD}~\cite{paad} achieves the strongest overall degradation due to full gradient access, such white-box information is not available in black-box threat models.

SEBA also achieves the lowest FID (81.7), indicating that its perturbed observations remain visually close to the clean states.

In terms of query cost, PA-AD and OPTIMAL require $2$M environment interactions and $2$M victim queries during training, whereas SEBA trains with a fixed budget of $80$K environment queries and $400$K victim queries and requires \emph{zero} victim access at attack time. 
This makes SEBA more practical in scenarios where interaction with either the environment or the victim policy is restricted or expensive.

Overall, these Atari results show that SEBA generalizes beyond continuous control and remains effective in pixel-based discrete-action settings.

\begin{table*}[t]
\centering
\caption{\textbf{Ablation study on SEBA components.}
Each variant removes one module: \textbf{(-D)} removes the discriminator, 
\textbf{(-Noise)} replaces noisy perturbed states in Stage~1 with clean states.
and \textbf{(-WM)} removes the world model.
Lower reward and FID indicate stronger attacks.}
\label{tab:ablation}
% \resizebox{\textwidth}{!}{
\begin{tabular}{lccccc}
\toprule
\textbf{Task (Reward $\downarrow$)} & \textbf{Clean} & \textbf{SEBA} & \textbf{(-D)} & \textbf{(-Noise)} & \textbf{(-WM)} \\
\midrule
Cheetah Run   & $859.26_{\pm35.00}$ & $1.61_{\pm3.97}$ & $3.15_{\pm5.24}$ & $19.25_{\pm9.31}$ & $0.18_{\pm0.04}$ \\
Walker Walk   & $944.28_{\pm42.26}$ & $35.74_{\pm6.67}$ & $22.64_{\pm8.13}$ & $118.11_{\pm27.64}$ & $36.01_{\pm10.54}$ \\
Walker Run    & $718.15_{\pm26.79}$ & $17.09_{\pm4.73}$ & $22.44_{\pm2.54}$ & $60.27_{\pm11.23}$ & $28.04_{\pm14.98}$ \\
Reacher Hard  & $870.90_{\pm90.43}$ & $0.30_{\pm0.90}$ & $0.33_{\pm0.14}$ & $12.98_{\pm9.21}$ & $0.00_{\pm0.00}$ \\
Hopper Stand  & $849.60_{\pm128.49}$ & $1.25_{\pm3.75}$ & $3.10_{\pm2.89}$ & $10.78_{\pm13.18}$ & $0.35_{\pm0.49}$ \\
\midrule
FID $\downarrow$                  & / & 62.43 & 97.18 & 60.72 & 63.98 \\
\midrule
Train Env (total) $\downarrow$   & / & 160K & 160K & 160K & 800K \\
Train Vic (total) $\downarrow$   & / & 800K & 800K & 800K & 800K \\
% Atk. Vic (per-step) $\downarrow$ & / & 0 & 0 & 0 & 0 \\
\bottomrule
\end{tabular}
% }
\end{table*}

\subsection{Ablation Study}
\label{sec:ablation}

To assess the contribution of each component in SEBA, we evaluate three ablated variants:
\textbf{(-D)} removes the discriminator in Stage~2;
\textbf{(-Noise)} replaces the perturbed states in Stage~1 with clean states (i.e., training uses $s_t$ rather than $s_t + \delta_t$);
and \textbf{(-WM)} removes the world model, 
training both the critic and the generator using only real environment rollouts.
All other training settings and hyperparameters are kept identical across variants.
Quantitative results are summarized in Table~\ref{tab:ablation}.

\paragraph{Discriminator.}
Removing the discriminator \textbf{(-D)} leads to a large increase in FID (from $62.43$ to $97.18$), showing that the perturbations are visually less consistent with clean observations. 
The change in reward is much smaller, meaning that the discriminator is not the main source of attack effectiveness. 
Its primary role is to stabilize the generator so that perturbations remain visually smooth, while the performance degradation is still mainly driven by the shadow critic.

\paragraph{Stage~1 Perturbed States.}
Replacing perturbed states with clean states \textbf{(-Noise)} yields the largest performance drop (e.g., Walker Walk: $35.74\!\to\!118.11$).
In Stage~1, the shadow critic is trained on perturbed states \(s'_t = s_t + G_\phi(s_t)\), so that the TD targets in Eq.~\eqref{eq:td1} reflect the cumulative rewards the agent obtains under perturbations.
This alignment ensures that \(Q_{\text{shadow}}\) provides meaningful gradients to guide \(G_\phi\).

When clean states are used instead, 
the critic is learned under a mismatched state distribution,
making its value estimates less reliable under perturbations,
which in turn leads to substantially weaker attacks in \textbf{(-Noise)}.

\paragraph{World Model.}
Removing the world model \textbf{(-WM)} yields comparable or slightly higher attack strength, but increases environment queries from $160\text{K}$ to $800\text{K}$.
The world model does not directly boost attack performance; instead, it enhances sample efficiency by generating synthetic transitions for updating both the shadow critic and the generator.

\paragraph{Summary.}
The three components serve complementary purposes:
the discriminator stabilizes visual appearance,
Stage~1 perturbed states ensure the critic remains reliable under the perturbed environment,
and the world model reduces environment query cost.
Together, these components yield a method that 
is both effective and efficient.

\begin{table*}[!t]
    \caption{\textbf{Targeted attack success rates (\%)}: an attack is counted as success if the selected action dimension $a_t^{(i)}$ falls within the target range $\mathcal{R}_{\text{target}}\subset[-1.0,1.0]$.}
    \label{tab:targeted_attack}
    \centering
    % \resizebox{\linewidth}{!}{
    \begin{tabular}{lccc|ccc}
        \hline
        \textbf{Task} & \textbf{Selected Dimension} & \textbf{Target Range} & & \textbf{PGD (\%) $\uparrow$} & \textbf{Critic-Based (\%) $\uparrow$} & \textbf{SEBA (\%) $\uparrow$} \\
        \hline
        Cheetah Run    & $4$ of $6$ & $[0.3,\ 0.5]$       & & 60.8 & 68.1 & \textbf{96.6} \\
        Walker Walk    & $5$ of $6$ & $[-0.6,\ -0.4]$     & & 37.2 & 53.2 & \textbf{91.3} \\
        Walker Run     & $2$ of $6$ & $[-0.7,\ -0.5]$     & & 58.0 & 64.5 & \textbf{95.6} \\
        Reacher Hard   & $2$ of $2$ & $[0.0,\ 0.2]$       & & 46.2 & 57.4 & \textbf{93.2} \\
        Hopper Stand   & $3$ of $4$ & $[0.6,\ 0.8]$       & & 76.6 & 84.5 & \textbf{98.8} \\
        \hline
    \end{tabular}
    % }
\end{table*}
\subsection{Targeted Attack Evaluation}
We evaluate whether SEBA can be adapted to achieve \emph{targeted} control.  
The victim's action space is $[-1.0,1.0]^m$; we therefore select a target sub-interval \(\mathcal{R}_{\text{target}}\subset[-1.0,1.0]\) for a single action dimension \(i\).  
An attack is counted as successful if the victim's executed action satisfies \(a_t^{(i)}\in\mathcal{R}_{\text{target}}\).  

\paragraph{Setup.}
To adapt SEBA for targeted control, we adopt the training structure in Eq.~\eqref{eq:gan_losses} and modify only the generator objective to encourage the victim's chosen action dimension \(i\) to fall within the target interval \(\mathcal{R}_{\text{target}}\subset[-1.0,1.0]\). 
The modified part of the generator loss is given by:
\begin{equation}
\max_{\phi}\; Q_{\text{shadow}}\!\big(s_t + G_\phi(s_t),\, a_t\big)
\quad \text{s.t.}\;\;
a_t^{(i)} \in \mathcal{R}_{\text{target}}.
\label{eq:targeted_obj}
\end{equation}

Here, \(a_t\) denotes the victim's action under the perturbed observation \(s_t + G_\phi(s_t)\).
This adjustment encourages the generator to produce perturbations that steer the selected action dimension into the target interval, while still maintaining visually subtle modifications consistent with the discriminator constraint.

\paragraph{Results and discussion.}
Table~\ref{tab:targeted_attack} shows that SEBA attains substantially higher targeted-success rates than PGD and the Critic-Based attacker across all tasks. For example, on \textit{Walker Walk} SEBA achieves \textbf{91.3\%} success versus 37.2\% (PGD) and 53.2\% (Critic-Based); on \textit{Reacher Hard} SEBA reaches \textbf{93.2\%} compared to 46.2\% and 57.4\%. These results indicate that SEBA can not only degrade overall performance 
but also steer specific action components toward desired target ranges.
This demonstrates that the perturbations are not only disruptive but can be shaped to enforce fine-grained control behavior when required.
\section{Conclusion}
\label{sec:conclusion}

This work presented \textbf{SEBA}, a sample-efficient framework for black-box adversarial attacks on visual reinforcement learning agents. 
By combining a shadow critic, a GAN-based perturbation generator, and a learned world model, SEBA enables strong attacks without access to gradients or internal model parameters. 
Experiments on pixel-based MuJoCo and Atari benchmarks show that SEBA achieves superior performance in terms of reward degradation, visual imperceptibility, and query efficiency, outperforming both white-box and black-box baselines. 
Moreover, SEBA extends naturally to targeted control tasks, demonstrating flexible and broadly applicable attack capability across visual RL domains. 
These results highlight SEBA as a practical and effective approach for evaluating the robustness of embodied AI systems under limited-access conditions.

% \clearpage
\section*{Acknowledgements}
This work was supported by the Ministry of Science and Technology of the People’s Republic of China (Grant No: 2025YFE0200100), the National Natural Science Foundation of China (Grant No: 62372122, 62502416, and U25A20430), and the Research Grants Council of Hong Kong SAR, China (Grant No: 15208923, and 15224124). The authors thank Di Wu and Yihan Zhang for their companionship and trust.
{
    \small
    \bibliographystyle{ieeenat_fullname}
    \bibliography{main}
}


\end{document}